\def\year{2020}\relax
\documentclass[letterpaper]{article} 
\usepackage{aaai20}  
\usepackage{times}  
\usepackage{helvet} 
\usepackage{courier}  
\usepackage[hyphens]{url}  
\usepackage{graphicx} 
\usepackage{graphicx}  
\usepackage{booktabs}
\usepackage[ruled]{algorithm2e}
\usepackage{bm}
\usepackage{amssymb}
\usepackage{amsmath}
\usepackage{array}
\usepackage{multirow}
\usepackage{subfigure} 
\title{Independence Promoted Graph Disentangled Networks}
\author{
Yanbei Liu\textsuperscript{\rm 1},
Xiao Wang\textsuperscript{\rm 2}\thanks{Corresponding author, Xiao Wang, xiaowang@bupt.edu.cn.},
Shu Wu\textsuperscript{\rm 3},
Zhitao Xiao\textsuperscript{\rm 1}\\
\textsuperscript{\rm 1}School of Life Sciences, Tiangong University \\
\textsuperscript{\rm 2}School of Computer Science, Beijing University of Posts and Telecommunications\\
\textsuperscript{\rm 3}Center for Research on Intelligent Perception and Computing, National Laboratory \\ of Pattern Recognition, Institute of Automation, Chinese Academy of Sciences\\
liuyanbei@tjpu.edu.cn, xiaowang@bupt.edu.cn, shu.wu@nlpr.ia.ac.cn, xiaozhitao@tjpu.edu.cn
}

\begin{document}

\maketitle

\begin{abstract}
We address the problem of disentangled representation learning with independent latent factors in graph convolutional networks (GCNs). The current methods usually learn node representation by describing its neighborhood as a perceptual whole in a holistic manner while ignoring the entanglement of the latent factors. However, a real-world graph is formed by the complex interaction of many latent factors (e.g., the same hobby, education or work in social network).
While little effort has been made toward exploring the disentangled representation in GCNs.
In this paper, we propose a novel Independence Promoted Graph Disentangled Networks (IPGDN) to learn disentangled node representation while enhancing the independence among node representations. 
In particular, we firstly present disentangled representation learning by neighborhood routing mechanism, and then employ the Hilbert-Schmidt Independence Criterion (HSIC) to enforce independence between the latent representations, which is effectively integrated into a graph convolutional framework as a regularizer at the output layer. Experimental studies on real-world graphs validate our model and demonstrate that our algorithms outperform the state-of-the-arts by a wide margin in different network applications, including semi-supervised graph classification, graph clustering and graph visualization.

\end{abstract}

\section{Introduction}
Graph with a set of objects and their relationships is a significant kind of data structure, which has been adopted in various areas such as social networks  \cite{hamilton2017inductive} and protein-protein interaction networks \cite{fout2017protein}.
Graph convolutional networks (GCNs), as a typical deep learning technique on graph data, have attracted considerable attentions \cite{defferrard2016convolutional,kipf2017semi-supervised}.
GCNs extend the traditional convolution operation to graph and it learns the node representation by propagating its neighbor information on graph.
With the learned node representations, we can perform various tasks on graphs such as node clustering, classification and link prediction \cite{zhang2018deep,wu2019comprehensive}

\begin{figure*}[htb]
\centering
\includegraphics[width=.95\textwidth]{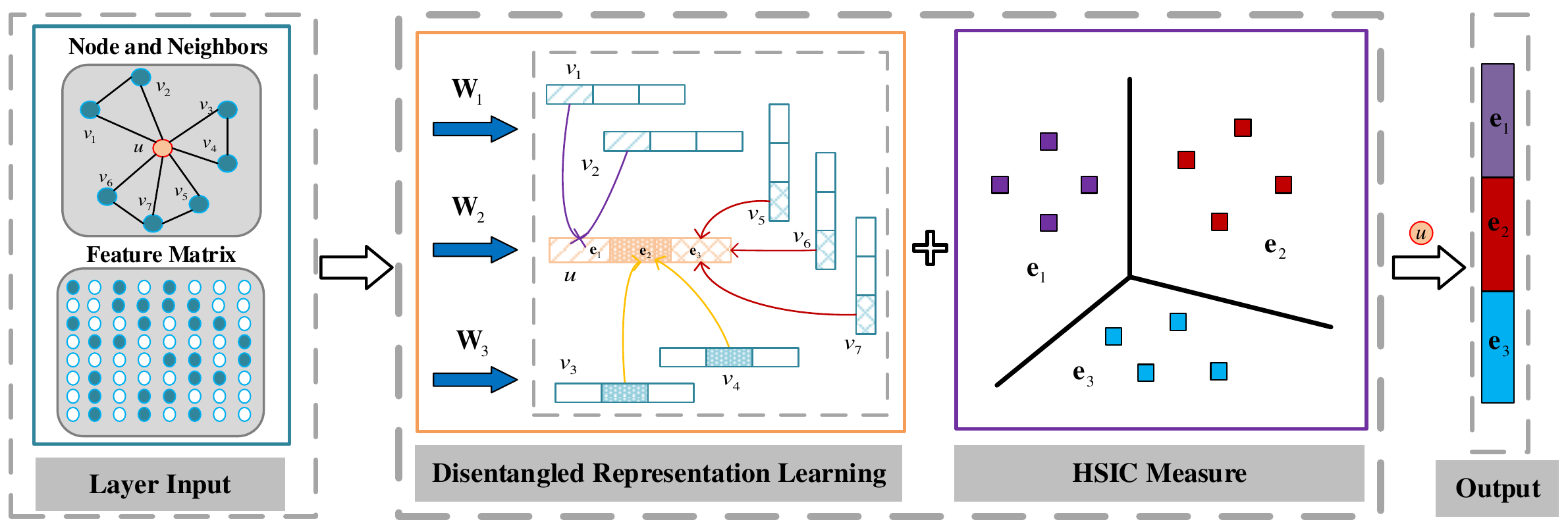}
\caption{Illustration of the proposed IPGDN's layer. It attempts to disentangle latent factors in the graph data, and simultaneously respects independence across different representations. To disentangle node $u$, IPGDN's layer first construct features from different aspects of its neighbors via disentangled representation learning, then encourage the independence among latent representations through minimize HSIC to obtain the final result. This example assumes that there are three latent factors, corresponding to the three channels $\mathbf{W}_1, \mathbf{W}_2, \mathbf{W}_3 $.}
\label{fig:model}
\end{figure*}

Despite the remarkable performance, the existing graph convolutional networks generally learn node representations by absorbing the node's neighborhood as a perceptual whole, and the representations they have learned are prone to over-smoothing faced by a deep GCNs \cite{li2018deeper}. 
Yet graphs are usually formed by highly complex interactions of many latent factors. 
For example, a person in a social network usually connects with others for various reasons (e.g., hobby, education and work), hence it simultaneously contains several partial information from its neighbors. 
Generally speaking, latent factors can take into account the nuances between the different parts of the neighborhood and make node representations more informative.
Furthermore, latent factors can enable to discover the underlying structures in graph data, such as neighborhood and community. Therefore, 
how to learn representations that capture desired latent factors behind a graph is of great importance for GCNs.

However, disentangling the latent factors behind a graph has two main challenges. One is how to learn node disentangled representations which need extract the different parts of its neighborhood but not the whole. 
Yet graph has numerous nodes and an unusually complex  network structure.
The other is how to encourage independence among the learned latent factors. Taking social networks as an example, although ``roommate" and ``classmate" are  two different factors that lead people to connect each other, the former is obviously a subset of the latter, hence they are not sufficiently diversified and will inevitably result in redundant representations. 
We argue that the representations of latent factors should be good and different. 
Yet the relationship between these factors is usually very complicated and nonlinear, and the solution needs to be differentiable so as to support end-to-end training in GCNs. So it is of very importance to encourage independence among the learned latent factors.
We notice that Ma et al. \cite{ma2019disentangled} presented a disentangled graph convolutional network to learn disentangled node representations. 
However, they only consider the disentangled representation learning while neglecting the independence of the latent factors. Therefore, how to learn representations that disentangle the latent factors with desired independence property remains largely unexplored  in the field of graph convolutional networks.

In this paper, we propose Independence Promoted Graph Disentangled Networks (IPGDN), a novel approach for disentangled representation learning that can automatically discover the independent latent factors in graph data. Specifically, on the one hand, we present disentangled representation learning by neighborhood routing mechanism. On the other hand, to enforce the independence among different latent representations, our model minimize the dependence among different representations with a kernel-based measure, in particular the Hilbert-Schmidt Independence Criterion (HSIC). 
The node disentangled representation learning and independence regularization are jointly optimized in a unified framework so that it finally leads to a better graph disentangled representation.
The experimental results on benchmark datasets demonstrate the superb performance of our approach on  graph analytic tasks, including semi-supervised node classification, node clustering, and graph visualization.

We summarize the main contributions as follows:

\begin{itemize}
\item To our best knowledge, this is the first attempt to enforce the independence property among different latent representations in graph disentangled representation learning. 
We propose a novel independently regularized disentanglement framework for graph convolutional network, 
which represents topological structure and node content into different latent factors. 
\item We present a kernel Hilbert-Schmidt Independence Criterion to effectively measure dependence among different latent representations, which can promote the quality of disentangled representations of the graph.
\item Experiments on benchmark graph datasets demonstrate that our graph disentangled networks outperform the others on three popular tasks.
\end{itemize}

\section{Graph Convolutional Networks}

First, let us define some notations used throughout this paper. Let $\mathcal{G} = (\mathcal{V,E},\mathbf{X})$ be an attributed information network, where $\mathcal{V}$ denotes the set of $n$ nodes and $\mathcal{E}$ represents the set of edges. $\mathbf{X} \in \mathbb{R}^{n\times f}$ is a matrix that encodes all node attribute information, and $\mathbf{x}_i$ describes the attribute associated with node $i$. We denote $(u,v)\in \mathcal{E}$ as an edge between nodes $u$ and $v$. Here we primarily focus on undirected graphs.

Graph convolutional networks generalize traditional convolutional neural networks to the graph domain.
Though a number of different GCNs methods have been proposed, here we focus on a representative one proposed by Kipf and Welling \cite{kipf2017semi-supervised}. Given a graph with adjacency matrix $\mathbf{A}$ and node feature matrix $\mathbf{X}$, its propagation rule is as follows:

\begin{equation}
\mathbf{H}^{(l)} = \rho \big( \widetilde{\mathbf{D}}^{-\frac{1}{2}} \widetilde{\mathbf{A}} \widetilde{\mathbf{D}}^{-\frac{1}{2}} \mathbf{H}^{(l-1)} \mathbf{W}^{(l-1)}  \big),
\end{equation}
where $\mathbf{A}$ is the adjacent matrix, $\widetilde{\mathbf{A}} = \mathbf{A} + \mathbf{I}$ and $\widetilde{\mathbf{D}}=\sum_j \widetilde{A}_{ij}$. $\mathbf{H}^{(l)}$ is the matrix of activations in the $l$-th layer, and $\mathbf{H}^{(0)}=\mathbf{X}$. 
 $\rho(\cdot)$ is a non-linear activation function such as ReLU and $\mathbf{W}^{(l-1)}$ are trainable parameters. 
The general philosophy is that nodes should exchange information with their immediate neighbors in each convolutional layer, followed by applying learnable filters and some non-linear transformation. 
This architecture can be trained end-to-end using task-specific loss function, for example, the cross entropy loss in semi-supervised node classification as follows:

\begin{equation}
\mathcal{L}_{cel} = - \sum_{v \in \mathcal{V}^{L}} \sum_{c=1}^C \mathbf{Y}_{v,c} log(\widetilde{ \mathbf{Y}}_{v,c}),
\label{eq:cel}
\end{equation}
where $\mathcal{V}^L$ is the set of labeled nodes, $C$ is the number of classes, $\mathbf{Y}$ is the label matrix and $\widetilde{ \mathbf{Y}} = softmax(\mathbf{H}^{(L)})$ are predictions of GCNs by passing the hidden representation in the final layer $\mathbf{H}^{(L)}$ to a softmax function.

\section{IPGDN: the Proposed Model}

This paper is concerned with independently disentangled representations for processing graph data. As shown in Figure \ref{fig:model}, we design our IPGDN's layer simultaneously integrating disentangled representation learning and independence among different representations into an unified framework. In this section, we firstly present a way of disentangled representation learning in graph. Secondly, we propose a measure of dependence across latent representations based on HSIC. Thirdly, we introduce our IPGDN architecture: the model to enhance independent representations  in semi-supervised classification task.

\subsection{Disentangled Representation Learning}

We assume that each node is composed of $M$ independent components, i.e., there are $M$ latent factors (corresponding to $\mathbf{M}$ channels) to be disentangled. Given nodes $\textbf{x}_u \in \mathbb{R}^{f}$ and $ \{\textbf{x}_v \in \mathbb{R}^{f}: (u,v)\in \mathcal{E} \} $ as input, we define $\textbf{h}_u = [\textbf{e}_1, \textbf{e}_2,...,\textbf{e}_M] \in  \mathbb{R}^{f_{l}}$ as the hidden representations of the node $u$, where $\textbf{e}_m \in \mathbb{R}^{\frac{f_{l}}{M}}(1 \leq m \leq M)$ is the $m$-th disentangled representation. 
The component $\mathbf{e}_m$ is for describing the aspect of node $u$ that are pertinent to the $m$-th factor.

For a node $o\in \{u \} \cup \{ v:(u,v)\in \mathcal{E} \}$, to extract its features from different aspects of its neighbors, we firstly project its feature vector $\textbf{x}_o$ into different subspaces as follows:
\begin{equation}
\textbf{z}_{o,m} = \rho(\textbf{W}^T_m \textbf{x}_o + \textbf{b}_m),
\label{eq:con}
\end{equation}
where $\textbf{W}_m \in \mathbb{R}^{f_{l-1} \times \frac{f_{l}}{M}}$ and $\textbf{b}_m \in \mathbb{R}^{\frac{f_{l}}{M}}$ are the parameters of $m$-th channel. To ensure numerical stability and prevent the neighbors with overly rich features, we use $\ell_2$ norm for normalization as $\mathbf{z}_{o,m}/\| \mathbf{z}_{o,m} \|_2$.  

To comprehensively capture aspect $m$ of node $u$, 
we need to construct $\textbf{e}_m $ from both $\textbf{z}_{u,m}$ and $\{ \textbf{z}_{v,m}:(u,v)\in \mathcal{E} \}$, which can identify the latent factor  and assign the neighbor to a channel.
Here we leverage neighborhood routing mechanism  \cite{ma2019disentangled} as follows:

\begin{equation}
\textbf{e}^{(t)}_{m} = \frac{\textbf{z}_{u,m} + \sum_{v:(u,v)\in \mathcal{E}} p^{(t-1)}_{v,m} \textbf{z}_{v,m} }{\left\| \textbf{z}_{u,m} + \sum_{v:(u,v)\in \mathcal{E}} p^{(t-1)}_{v,m} \textbf{z}_{v,m} \right\|_2}, 
\label{eq:em}
\end{equation}

\begin{equation}
p^{(t)}_{v,m} = \frac{exp(\textbf{z}_{v,m}^T \textbf{e}^{(t)}_{m})}{\sum^{M}_{m=1} exp(\textbf{z}_{v,m}^T \textbf{e}^{(t)}_{m})},
\label{eq:pvm}
\end{equation}
where iteration $t = 1,2,...,T$. $p_{v,m}$ indicates the probability that factor $m$ is the reason why node $u$ reaches neighbor $v$, and satisfies $p_{v,m} \geq 0, \sum^{M}_{m=1} p_{v,m}=1$. Also, $p_{v,m}$ is the probability that we should use neighbor $v$ to construct $\textbf{e}_m$. The neighborhood routing mechanism will iteratively infer $p_{v,m}$ and construct $\textbf{e}_m$.

\subsection{Independence across Latent Representations}

To enhance the disentangled informativeness, we encourage the representations of different factors to be of sufficient independence by using HSIC.

For latent factors $1 \leq i,j \leq M, i \neq j$, let $\mathbf{e}_i$ and $\mathbf{e}_j$ be disentangled representations from two latent factors containing $\frac{f_{l}}{M}$ data points $\{ (e_{i,p}, e_{j,p}) \in \mathcal{X} \times \mathcal{Y}  \}_p^{\frac{f_{l}}{M}}$ , which are jointly drawn from a probability distribution $P_{\mathbf{e}_i,\mathbf{e}_j}$. 
$\phi(\mathbf{e}_i)$ and $\psi(\mathbf{e}_j)$ are functions that map $\mathbf{e}_i \in \mathcal{X}$ and $\mathbf{e}_j \in \mathcal{Y}$ to kernel space $\mathcal{F}$ and $\Omega$ with respect to the kernel functions $k(e_{i,p},e_{i,q})= <(\phi(e_{i,p}),\phi(e_{i,q})>$ and $s(e_{j,p},e_{j,q})= <(\phi(e_{j,p}),\phi(e_{j,q})>$. Note that $\mathcal{F}$ and $\Omega$ are Reproducing Kernel Hilbert Space (RKHS) on $\mathcal{X}$ and $\mathcal{Y}$, respectively.
The cross-covariance is a function that gives the covariance of two random variables and defined as follows: 
\begin{equation}
\mathcal{C}_{\mathbf{e}_i,\mathbf{e}_j} = E_{\mathbf{e}_i,\mathbf{e}_j}[(\phi(\mathbf{e}_i) - \mu_{\mathbf{e}_i}) \otimes (\psi(\mathbf{e}_j) - \mu_{\mathbf{e}_j})],
\end{equation}
where $\otimes$ is the tensor product. Then, HSIC as the Hilbert-Schmidt norm of the associated cross-covariance operator $\mathcal{C}_{\mathbf{e}_i,\mathbf{e}_j}$ is defined as follows:
\begin{equation}
\text{HSIC}(P_{\mathbf{e}_i,\mathbf{e}_j},\mathcal{F},\Omega) := \left\| \mathcal{C}_{\mathbf{e}_i,\mathbf{e}_j} \right\|^2_{HS},
\end{equation}
where $\left\| \mathbf{A} \right\|_{HS} = \sqrt{\Sigma_{i,j}a^2_{ij}}$.

Accordingly, the empirical version of HSIC \cite{gretton2005measuring} is given as follows:
\begin{equation}
\text{HSIC}(\mathbf{e}_i,\mathbf{e}_j) = (\frac{f_{l}}{M} -1)^{-2}tr(\mathbf{K}\mathbf{R}\mathbf{S}\mathbf{R}),
\label{eq:hsic}
\end{equation}
where $\mathbf{K}$ and $\mathbf{S}$ are the Gram matrices with $k_{p,q} = k(e_{i,p},e_{i,q}),s_{p,q} = s(e_{j,p},e_{j,q})$. $r_{i,j} = \delta_{i,j} - \frac{M}{f_{l}}$ centers the
Gram matrix to have zero mean in the feature space. 
There are two main advantages to use the HSIC to measure the dependence of representations. First, HSIC maps representations into a reproducing kernel Hilbert space to measure their dependence such that correlations measured in that space correspond to high-order joint moments between the original distributions and more complicated (such as nonlinear) dependence can be addressed. 
Second, this method is able to estimate dependence between representations without explicitly estimating the joint distribution of the random variables.
In our implementation, we use the inner product kernel function, i.e. $\mathbf{K} = \mathbf{e}_i \mathbf{e}_i^T$, and promising performances are achieved. Note that minimizing $\text{HSIC}(\mathbf{e}_i,\mathbf{e}_j)$ enhances the independence between $\mathbf{K}$ and $\mathbf{S}$, which penalties the consistency between kernel matrices from different latent representation parameterized by different projection matrices $\mathbf{W}$. 

\subsection{Network Architecture}

\begin{algorithm}[htb]
\footnotesize
\caption{The Proposed IPGDN's Layer}
\KwIn{ $\mathbf{x}_u \in \mathbb{R}^{f_{l-1}}$:the feature vector of node $u$; \\
\quad $\{\mathbf{x}_v \in \mathbb{R}^{f_{l-1}}:(u,v)\in  \mathcal{G} \}$:its neighbors' features; \\
\quad $M$:number of channels;  \\
\quad $T$:iterations of routing.
     }
\KwOut{$\mathbf{h}_u \in \mathbb{R}^{f_{l}}$: node disentangled representations.}
\textbf{Param}: $\mathbf{W}_m \in \mathbb{R}^{f_{l-1}\times \frac{f_l}{M} }$,\quad $\mathbf{b}_m \in \mathbb{R}^{\frac{f_l}{M}}$.\\

 \While{each $o \in \{u \} \cup \{v:(u,v)\in \mathcal{G} \}$}
 {
     \For{each $m \in M$}
     {
       Calculate $\mathbf{z}_{o,m}$ by Eq. (\ref{eq:con}); \\
        $\mathbf{z}_{o,m}\longleftarrow  \mathbf{z}_{o,m}/\| \mathbf{z}_{o,m} \|_2 $.
     }
 }
 
 \While{routing iteration $t \in T$}
 {
     \For{each $v \in (u,v)\in  \mathcal{G}$}
     {
       Calculate $p_{v,m}^{(t)}$ by Eq. (\ref{eq:pvm}). 
     }
     \For{each $m \in M$}
     {
       Update $\mathbf{e}_{m}^{(t)}$ by Eq. (\ref{eq:em}).        
     }   
     
 }
 
  \While{latent factors $(i,j) \in M$}
 {
    Calculate dependence among representations by Eq. (\ref{eq:hsic});\\
    Minimize Eq. (\ref{eq:lreg}).
 }
 $\mathbf{h}_u \longleftarrow $ the concatenation of $\textbf{e}_1, \textbf{e}_2,...,\textbf{e}_M$.
 \label{al:ipgdn}
\end{algorithm}

In this section, we describe the overall network architecture of IPGDN for performing node-related tasks.

By using disentangled representation learning module, we design the propagation of the first $L$ layers' output as follows:
\begin{equation}
\mathbf{h}_u^{(l)} = \text{dropout}\big (h^{(l)}(\mathbf{h}_u^{(l-1)},\{ \mathbf{h}_v^{(l-1)}:(u,v)\in \mathcal{E} \})\big ),
\end{equation}
where $h^{(l)}$ denotes the output function of the $l$-th layer, $1 \leq l \leq L$. $u\in \mathcal{V}$ and $\mathbf{h}_u^{(0)}=\mathbf{x}_u $. The dropout operation is appended after every layer and is enabled only during training. 

To enhance that the learned representations are independent, we incorporate the HSIC regularization at the $L$-th output layer $\textbf{h}_u^{(L)} = [\textbf{e}_1^{(L)}, \textbf{e}_2^{(L)},...,\textbf{e}_M^{(L)}]$, which is shown as follows:
\begin{equation}
\mathcal{L}_{reg} =  \sum_{u\in \mathcal{V}^L } \sum_{\mathbf{e}_i \neq \mathbf{e}_j} \text{HSIC}(\mathbf{e}_i^{(L)}, \mathbf{e}_j^{(L)}),
\label{eq:lreg}
\end{equation}

For semi-supervised classification task, the final layer of the network architecture is a fully-connected layer as follows:
\begin{equation}
\mathbf{h}^{(L+1)}=(\mathbf{W}^{(L+1)})^T \mathbf{h}^{(L)} + \mathbf{b}^{(L+1)}, 
\end{equation}
where transformation matrix $\mathbf{W}^{(L+1)} \in \mathbb{R}^{f_{L} \times C}$, parameters vector $\mathbf{b}^{(L+1)}\in \mathbb{R}^C$.

Next, $\mathbf{h}^{(L+1)}$ is passed to a softmax function to get the predicted labels:
\begin{equation}
\mathbf{\widetilde{y}} = softmax(\mathbf{h}^{(L+1)}).
\end{equation}

Then, we can use cross entropy loss $\mathcal{L}_{cel}$ defined in Eq. (\ref{eq:cel}) as an objective function.
Finally, we jointly minimize the loss function by combining the above terms:
\begin{equation}
\mathcal{L} = \mathcal{L}_{cel} + \lambda \mathcal{L}_{reg}
\end{equation}
where $\lambda$ is hyper-parameters that control the impact of different regularizations.
We compute the gradients via back-propagation, and optimize the parameters with Adam \cite{kingma2014adam}. Parameter $\lambda \geq 0$ is a trade-off parameter. The overall process of IPGDN's layer is shown in Algorithm \ref{al:ipgdn}. 

To summarize, our approach has the following merits: (1) The HSIC penalty is only based on the learned representation, i.e. network parameters--not additional assumptions. (2) Both disentangled representation learning and independence of different representations are addressed in an unified framework; (3) Our approach is easy to optimize with the Adam, and since the value of our objective function converges fast, the algorithm is effective for graph disentangled representation.

\subsection{Complexity Analysis for IPGDN Algorithm}

In our model, operation of the IPGDN's layer can be parallelized across all nodes. Thus, it is highly efficient. The computation complexity of our method is $\mathcal{O}\Big( |\mathcal{E}|  \sum_{l=0}^L f^{(l)} + |\mathcal{V}| (\sum_{l=1}^L f^{(l-1)}f^{(l)} + (f^{(L)})^2 ) + T \sum_{l=1}^L f^{(l)} \Big)$, where $|\mathcal{E}|$ is the number of edges, $|\mathcal{V}|$ is the number of nodes, $T$ is the iteration times used in neighborhood routing mechanism and $f^{(l)}$ is the dimensionality of the $l$-th hidden layer. 
It is worth noting that our algorithm is linear with respect to the number of nodes and number of edges in the graph respectively, which is in the same order as other GCNs.

\subsection{Independence Analysis of HSIC}

In this section, we analyze theoretical properties of the Hilbert-Schmidt Independence Criterion for enhancing independence among different representations.
Whenever $\mathcal{F}$, $\Omega$ are RKHS with characteristic kernels $k, s$ (in the sense of \cite{fukumizu2008kernel}), then $\text{HSIC}(P_{\mathbf{e}_i,\mathbf{e}_j},\mathcal{F},\Omega)=0$ if and only if $\mathbf{e}_i$ and $\mathbf{e}_j$ are independent. Intuitively, a characteristic kernel leads to an injective embedding of probability measures into the corresponding RKHS. The HSIC is the squared RKHS distance between the embedded joint distribution and the embedded product of the marginals. Examples of characteristic kernels are Gaussian RBF kernels and Laplace kernels. HSIC is zero if two representations are independent.
Note that non-characteristic and non-universal kernels can also be used for HSIC, although they may not guarantee that all dependence is detected. Different kernels incorporate distinctive prior knowledge into the dependence estimation, and they focus HSIC on dependence of a certain type. For instance, a linear kernel requires HSIC to seek only second order dependence, whereas a
polynomial kernel restricts HSIC to test for dependences of its degree. 
In terms of disentangled representation, to explore different latent factors behind a graph, we try to select independent representations that minimize HSIC by the inner product kernel. 

\section{Experiments}
In this section, we empirically perform comparative evaluation of IPGDN model against a wide variety of state-of-the-arts on three node-related tasks: semi-supervise node classification, node clustering and graph visualization. 

\begin{table}[tb]
\caption{Dataset statistics}
\centering
\begin{tabular}{lccc}  
\toprule
\textbf{Dataset} & \textbf{Cora} & \textbf{Citeseer} & \textbf{Pubmed} \\
\midrule
Nodes    & 2708 & 3327 & 19717        \\
Edges    & 5429 & 4732 & 44338         \\
Classes  & 7    & 6    & 3             \\
Features & 1433 & 3703 & 500           \\
Training Nodes   & 140 & 120 & 60        \\
Validation Nodes & 500 & 500 & 500       \\
Test Nodes       & 1000 & 1000 & 1000      \\
\bottomrule
\end{tabular}
\label{tab:dataset}
\end{table}

\subsection{Experimental Setup}

\textbf{Datasets}
We conduct our experiments on three standard citation network benchmark datasets, whose statistics are listed in Table \ref{tab:dataset}. Cora, Citeseer and Pubmed \cite{sen2008collective} are all for semi-supervised node classification and node clustering. We follow the experimental setup of \cite{yang2016revisiting}.
In all of these datasets, nodes correspond to documents and edges to (undirected) citations. Node features correspond to elements of a bag-of-words representation of a document. Each node has a class label, i.e., research area. We allow for only 20 nodes per class to be used for training. The predictive power of the trained models is evaluated on 1000 test nodes, and we use 500 additional nodes for validation purposes (the same ones as used by \cite{kipf2017semi-supervised,velivckovic2017graph}).

\begin{table}[!t]
\caption{Semi-supervised classification results (\%).}
\centering
\begin{tabular}{llccc}
\toprule
\multirow{2}*{\textbf{Method}} & \multirow{2}*{\textbf{Metrics}} & \multicolumn{3}{c}{\textbf{Datasets}} \\\cline{3-5}
          & & \textbf{Cora} & \textbf{Citeseer} & \textbf{Pubmed}  \\
\hline
MLP             &\multirow{13}*{ACC}     & 55.1  & 46.5 & 71.4  \\
                               
ManiReg         &     & 59.5  & 60.1 & 70.7  \\
                              
SemiEmb         &     & 59.0  & 59.6 & 71.1  \\
                              
LP              &     & 68.0  & 45.3 & 63.0  \\
                               
DeepWalk        &     & 67.2  & 43.2 & 65.3  \\
                               
ICA             &     & 75.1  & 69.1 & 73.9  \\
                               
Planetoid       &     & 75.7  & 64.7 & 77.2  \\
                               
ChebNet         &     & 81.2  & 69.5 & 75.0   \\
                               
GCN             &     & 81.4  & 71.2 & 79.0  \\
                               
MoNet           &     & 81.7  & -    & 78.8  \\
                               
GAT             &     & 83.0  & 72.9 & 78.7  \\
                               
DisenGCN        &     & 83.7  & 73.2 & 80.5   \\
                                
IPGDN (our)      &     & \textbf{84.1} & \textbf{74.0} & \textbf{81.2}\\ 
\hline
MLP             &\multirow{13}*{F1}      & 54.3  & 45.6 & 69.4 \\
ManiReg         &      & 60.2  & 61.2 & 70.2 \\
SemiEmb         &      & 57.9  & 59.1 & 70.0 \\
LP              &      & 66.2  & 46.2 & 62.8 \\
DeepWalk        &      & 70.3  & 51.5 & 66.1  \\
ICA             &      & 74.8  & 70.2 & 74.3\\
Planetoid       &      & 73.6  & 63.5 & 77.0 \\
ChebNet         &      & 80.9  & 66.2 & 75.5 \\ 
GCN             &      &81.2   & 71.1 & 78.8 \\
MoNet           &      & 81.5  &-     & 77.9 \\ 
GAT             &      & 82.5  & 71.4 & 77.7  \\   
DisenGCN        &      & 83.6  & 72.0 & 80.9 \\
IPGDN (our)      &      & \textbf{84.2}  & \textbf{72.8} & \textbf{81.7} \\
\bottomrule
\end{tabular}
\label{tab:semi_classi}
\end{table}

\textbf{Baselines}
We compare with some state-of-the-art baselines, including node embedding methods and graph neural network based methods, to verify the effectiveness of the proposed IPGDN.  

$\bullet$ MLP: It is a multi-layer perception as a baseline.

$\bullet$ ManiReg \cite{belkin2006manifold}: It is a semi-supervised learning model base on manifold regularization which allows to exploit the geometry of the marginal distribution.

$\bullet$ SemiEmb \cite{weston2012deep}:It is a semi-supervised embedding learning model incorporating deep learning techniques.

$\bullet$ LP \cite{zhu2003semi} : It is a label propagation approach based on a Gaussian random field model.

$\bullet$ DeepWalk \cite{perozzi2014deepwalk}: A random walk based network embedding method for graph.

$\bullet$ ICA \cite{lu2003link}: It is a link-based classification method which supports discriminative models describing both the link distributions and the attributes of linked objects.

$\bullet$ Planetoid \cite{yang2016revisiting}: This model is an inductive, embedding based approach to semi-supervised learning. It uses the graph structure as a form of regularization during training while not using graph structural information during inference.

$\bullet$ ChebNet \cite{defferrard2016convolutional}: It is a spectral graph convolutional network by means of a Chebyshev expansion of the graph Laplacian, removing the need to compute the eigenvectors of the Laplacian and yielding spatially localized filters.  

$\bullet$ GCN \cite{kipf2017semi-supervised}: It is a simple yet effective ChebNet model by restricting the filters to operate in a 1-step neighborhood around each node. 

$\bullet$ MoNet \cite{monti2017geometric}: This model is an extended CNN architectures by learning local, stationary, and compositional task-specific features for non-Euclidean data.

$\bullet$ GAT \cite{velivckovic2017graph}: It enhances GCN by introducing multi-head self-attention to assign different weights to different neighbors.

$\bullet$ DisenGCN \cite{ma2019disentangled}: This model is a graph convolutional network which tries to disentangle the latent factors among the complex graph by a neighborhood routing mechanism.

\textbf{Implementation Details}
In semi-supervised classification tasks, for a simple completion, we set the output dimension of DisenGCN and IPGDN's layer is a constant, i.e. $M \times \triangle f$ is the same in each layer, where $M$ is the number of channels used by the layer and $\triangle f $ is the output dimension of each channel. In node clustering task, we follow GAT and set the output dimension of a graph neural network to be 64. For DeepWalk method, we set window size to 5, walk length to 100, walks per node to 40 and the number of negative samples to 5. 
In our model, we use $K = 4, \bigtriangleup f = 16$ for the test. Following \cite{ma2019disentangled}, we set iterations of neighbourhood routing $T = 7 $. We then tune the hyper-parameters of both our model's and our baselines' automatically using hyperopt \cite{bergstra2013making}. Specifically, we run hyperopt for 200 trials for each setting, with the hyper-parameter search space specified as follows: the learning rate $\backsim$ loguniform $[e^{-8},1]$, the $\ell_2$ regularization term $\backsim$ loguniform$[e^{-10},1]$, dropout rate $\in \{ 0.05, 0.10,..., 0.95 \}$, the number of layers $L \in \{1, 2,..., 6 \} $. 
Then with the best hyper-parameters on the validation sets, we report the averaged performance of 50 runs on each semi-supervised dataset, and 30 runs in node clustering task.

\subsection{Classification Analysis}

\begin{table*}[!t]
\caption{Node clustering results(\%).}
\centering
\begin{tabular}{clcccccccc}  
\toprule
\textbf{Datasets} & \textbf{Metrics}  & \textbf{SemiEmb} & \textbf{DeepWalk} & \textbf{Planetoid} & \textbf{ChebNet} & \textbf{GCN}  & \textbf{GAT} & \textbf{DisenGCN}& \textbf{IPGDN}\\
\midrule
\multirow{5}*{Cora}     & ACC & 60.8 & 62.2  & 67.2 & 71.9 & 73.5  & 75.2  & 75.5  & \textbf{76.1}    \\
                        & NMI & 48.7 & 50.3  & 52.0 & 49.8 & 51.7  & 57.0  & 58.4  & \textbf{59.2}    \\ 
                        & ARI       &41.5 &40.8 &40.5 &42.4 &48.9 &54.1 &60.4 & \textbf{61.0}\\
                        & Precision &21.6 &22.6 &24.0 &27.6 &28.5 &23.7 &27.1 & \textbf{28.1} \\
                        & F1        &24.3 &21.8 &25.0 &25.9 &25.6 &23.6 &24.8 & \textbf{25.3}  \\ \hline
\multirow{5}*{Citeseer} & ACC & 51.1 & 52.1  & 61.0 & 65.0 & 67.7  & 68.0  & 68.2  & \textbf{68.9}    \\
                        & NMI & 31.2 & 30.5  & 41.2 & 42.6 & 42.8  & 43.1  & 43.7  & \textbf{44.3}    \\ 
                        & ARI       &21.5 &20.6 &22.1 &41.5 &42.8 &43.6 &42.5 & \textbf{43.0}\\
                        & Precision &22.9 &23.6 &20.9 &21.6 &20.5 &21.1 &23.8 & \textbf{23.9} \\
                        & F1        &20.0 &17.0 &19.5 &20.8 &19.9 &20.3 &22.6 & \textbf{23.2}  \\ \hline
\multirow{5}*{Pubmed}   & ACC & 62.3 & 65.2  & 64.6 & 75.2 & 75.6  & 76.3  & 77.0  & \textbf{77.8}   \\
                        & NMI & 27.8 & 29.6  & 32.5 & 35.6 & 35.0  & 35.0  & 36.1  & \textbf{37.0}   \\
                        & ARI       &35.2 &36.6 &33.9 &38.6 &40.9 &41.4 &41.6 &\textbf{42.0} \\
                        & Precision &21.3 &28.3 &30.0 &29.0 &32.5 &34.5 &33.6 &\textbf{34.0}  \\
                        & F1        &26.1 &25.9 &28.9 &31.3 &29.8 &30.3 &31.2 &\textbf{32.1}   \\ 
\bottomrule
\end{tabular}
\label{tab:node_clustering}
\end{table*}

We report the results in Table \ref{tab:semi_classi}.
From the Table \ref{tab:semi_classi}, we can see that IPGDN achieves the best performance in term of classification Accuracy (ACC) and F-score (F1).
More specifically, on the one hand, disentangled approaches, i.e. both DisenGCN and our model, outperform holistic approaches such as ChebNet, GCN and GAT, which indicates that the disentangled feature is helpful for graph representation.
On the other hand, we are able to improve upon DisenGCN by a margin of 1.7\%, 0.9\% and 1.7\% on Cora, Citeseer and Pubmed respectively, suggesting that encouraging independence between latent factors may be beneficial.

\subsection{Clustering Analysis}

To further evaluate the embeddings learned from the above algorithms, we also conduct the clustering task. For all the compared algorithms, we obtain its node embedding via feed forward when the model is trained. Then we use the KMeans to perform node clustering and the number of clusters $K$ is set to the number of classes. The same ground-truth as in node classification analysis is utilized. 
Following \cite{pan2018adversarially}, we employ five metrics to validate the clustering results: Accuracy (ACC), Normalized Mutual Information (NMI), Precision, F-score (F1) and Average Rand Index (ARI).
Since the performance of KMeans is affected by initial centroids, we repeat the process for 20 times and report the average results in Table \ref{tab:node_clustering}.

As can be seen in Table \ref{tab:node_clustering}, we can see that IPGDN performs consistently much better than all baselines. Also, graph neural network based algorithms usually achieve better performance. Besides, with the constraint of independence, IPGDN performs significantly better than DisenGCN . It shows that the proposed IPGDN can learn a more meaningful node embedding via enforcing independence between latent representations.

\subsection{Visualization}

To present a better intuitively comparation,  we provide a visualization of the $t$-SNE \cite{maaten2008visualizing} transformed feature representations. Specifically, we learn the node embedding based on the proposed model and project the learned embedding into a 2-dimensional space. Here we utilize $t$-SNE  to visualize the author embedding on Cora dataset and coloured the nodes based on their research areas.

\begin{figure*}[htb]
\centering
 \subfigure[GCN]{
 \includegraphics[width=.23\textwidth]{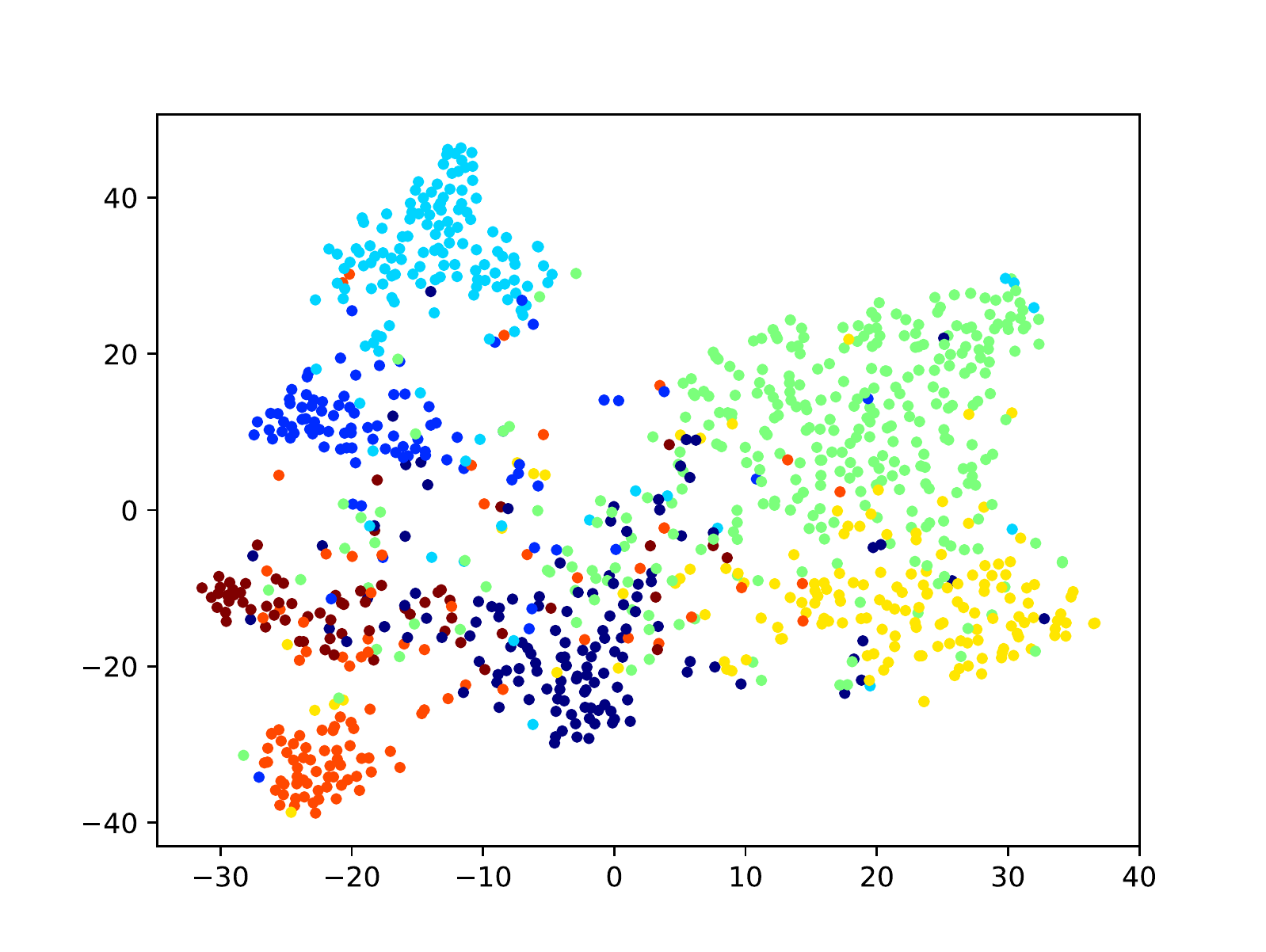}} 
 \subfigure[GAT]{
 \includegraphics[width=.23\textwidth]{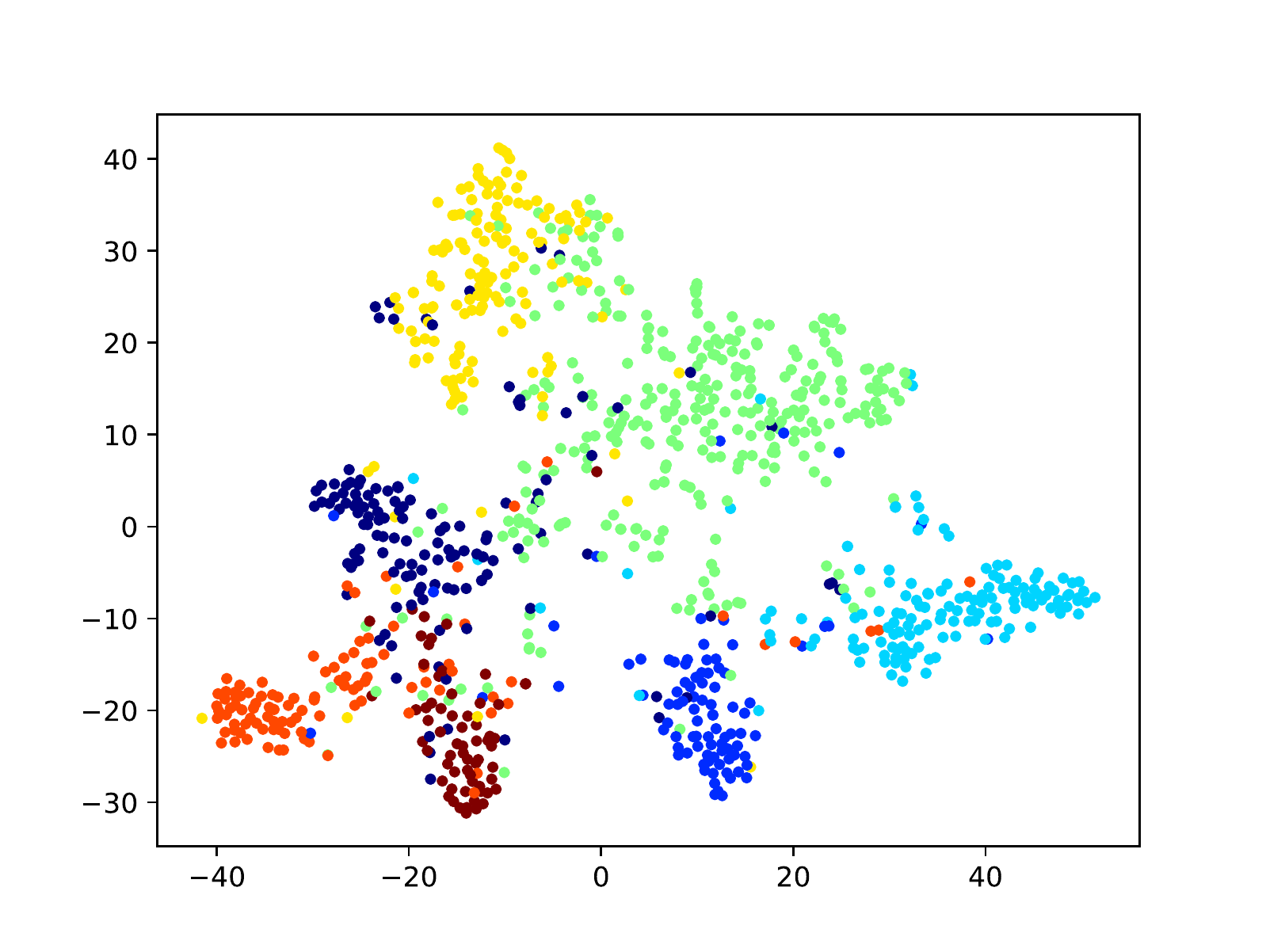}} 
 \subfigure[DisenGCN]{
 \includegraphics[width=.23\textwidth]{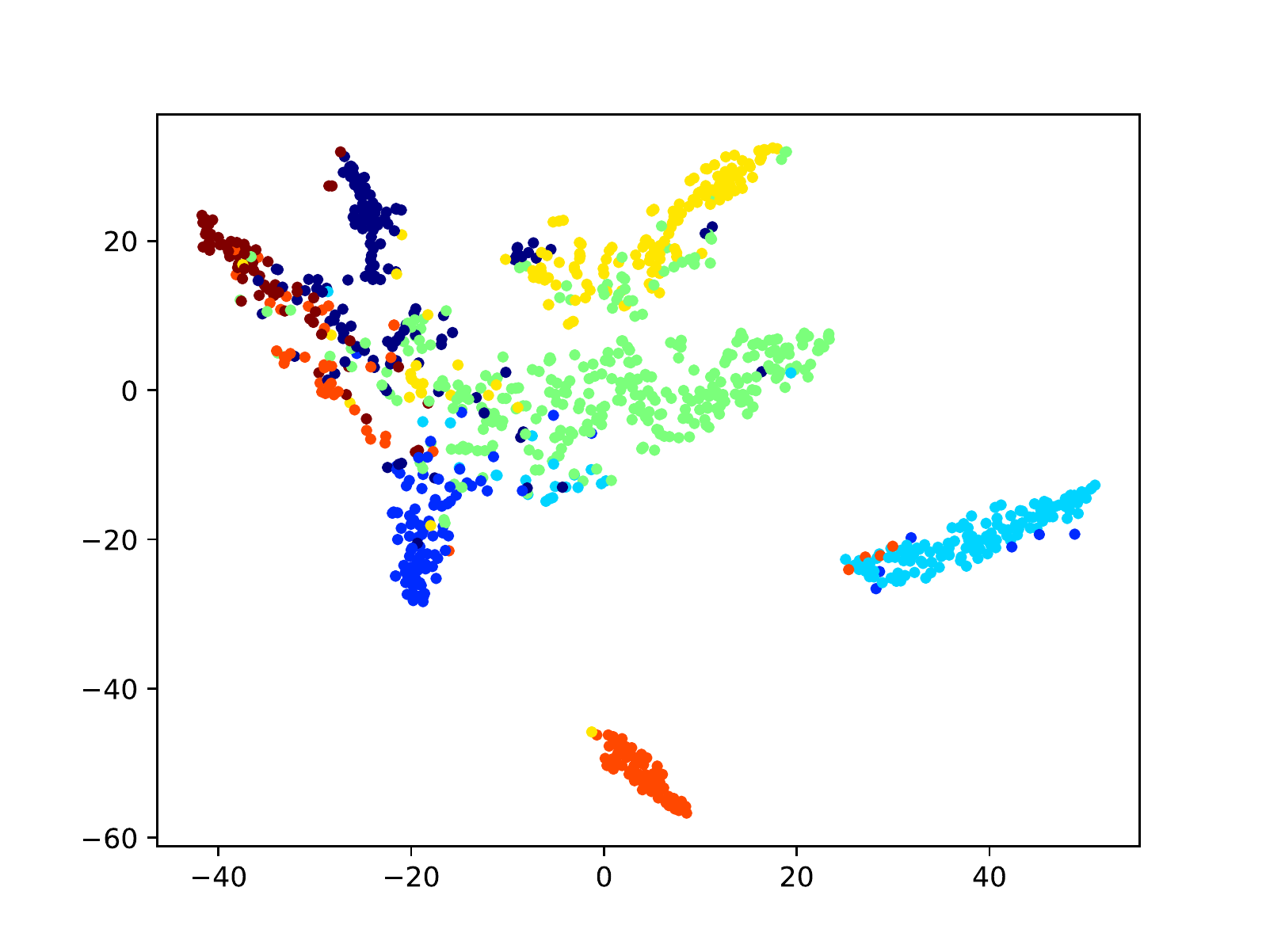}} 
 \subfigure[IPGDN]{
 \includegraphics[width=.23\textwidth]{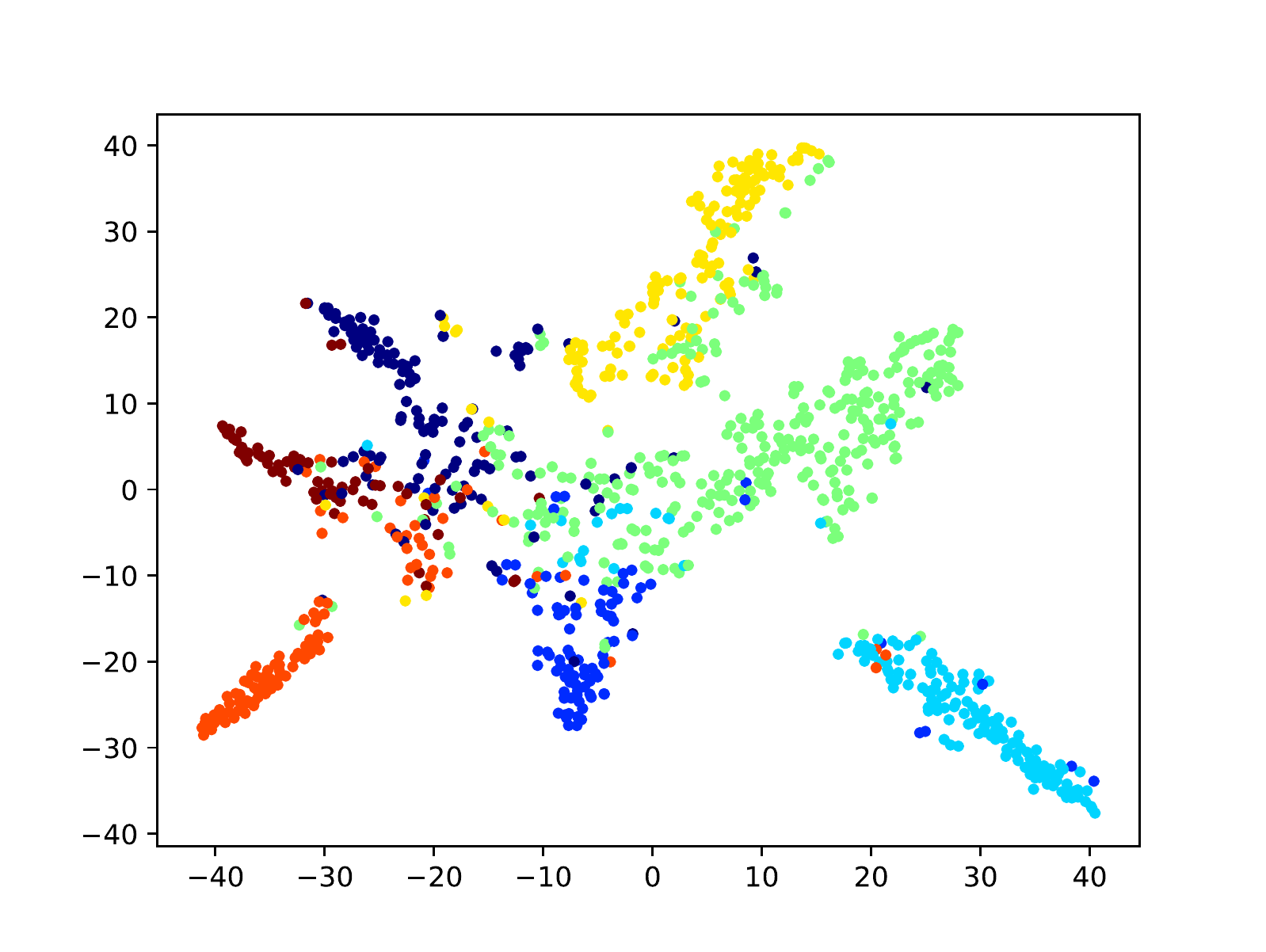}} 
 \caption{Visualization embedding on Cora. Each point indicates one author and its color indicates the research area. We can see that the embedding learned by IPGDN has high intra-class similarity and separates articles in different research area with distinct boundaries by encouraging independence among different latent representations.}
 \label{fig:vis}
 \end{figure*}

Base on Figure \ref{fig:vis}, we can see that  IPGDN performs much better than other graph neural networks and slightly better than DisenGCN. It demonstrates that the embedding learned by IPGDN has high intra-class similarity and separates the article in different research area with distinct boundaries by encouraging independence among different latent representations. 
On the contrary, GCN and GAT which get holistic feature from their neighbors do not perform well. 
The authors belong to different research areas are mixed with each other.

\subsection{Hyperparameter Sensitivity}

\begin{figure}[htb]
\centering
 \subfigure[Cora]{
 \includegraphics[width=0.31\columnwidth]{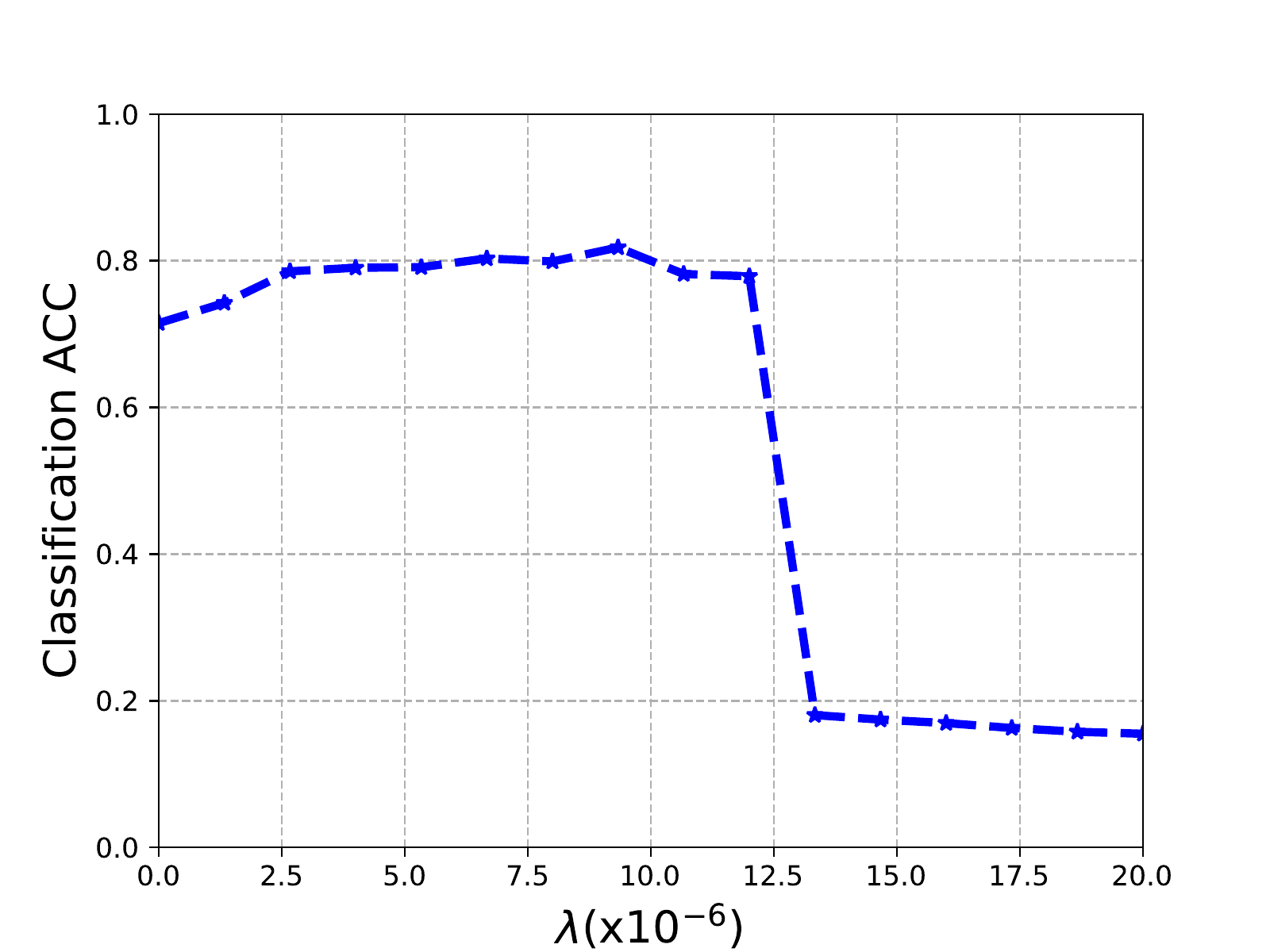}}
 \subfigure[Citeseer]{
 \includegraphics[width=0.31\columnwidth]{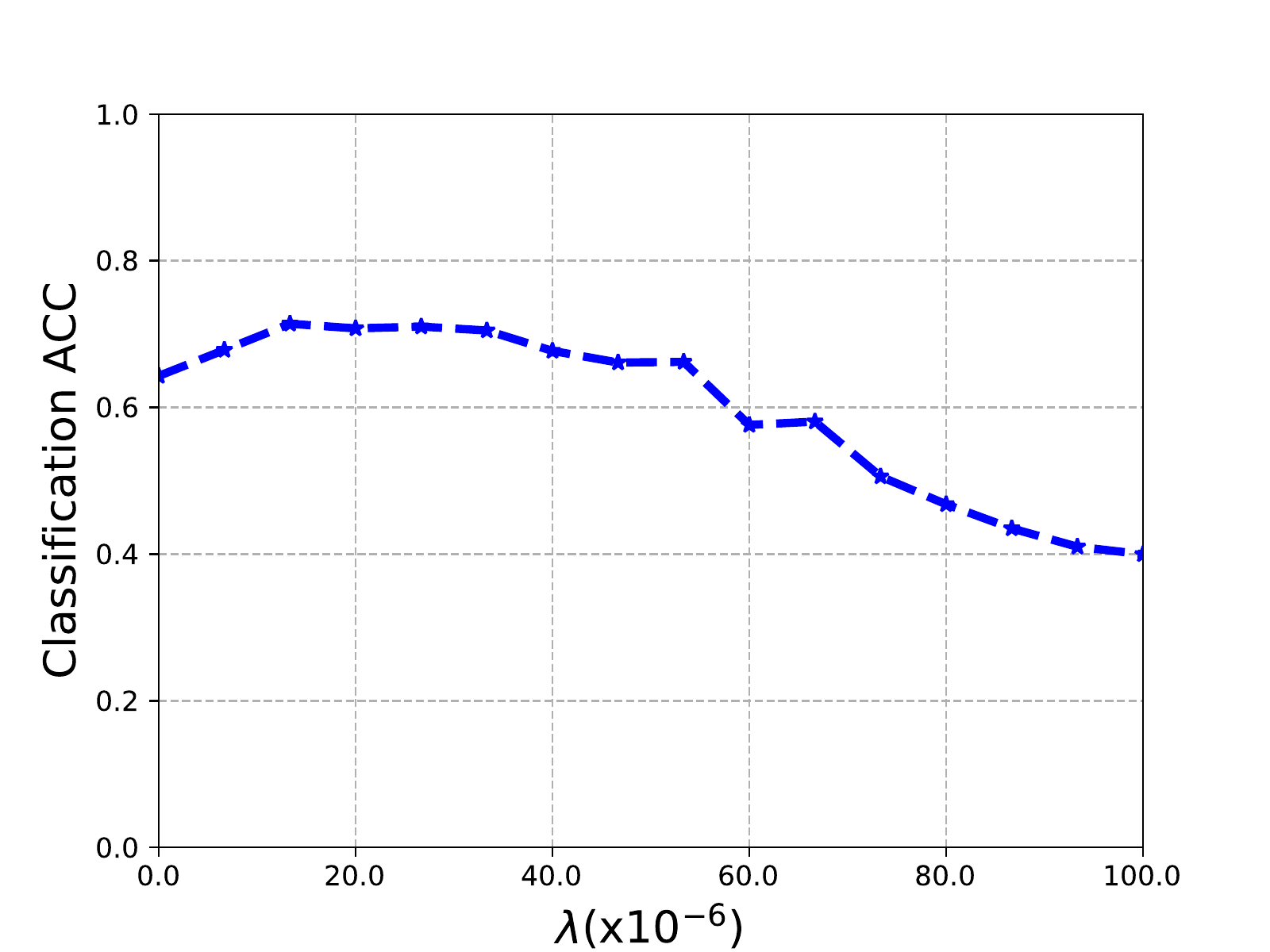}}
 \subfigure[Pubmed]{
 \includegraphics[width=0.31\columnwidth]{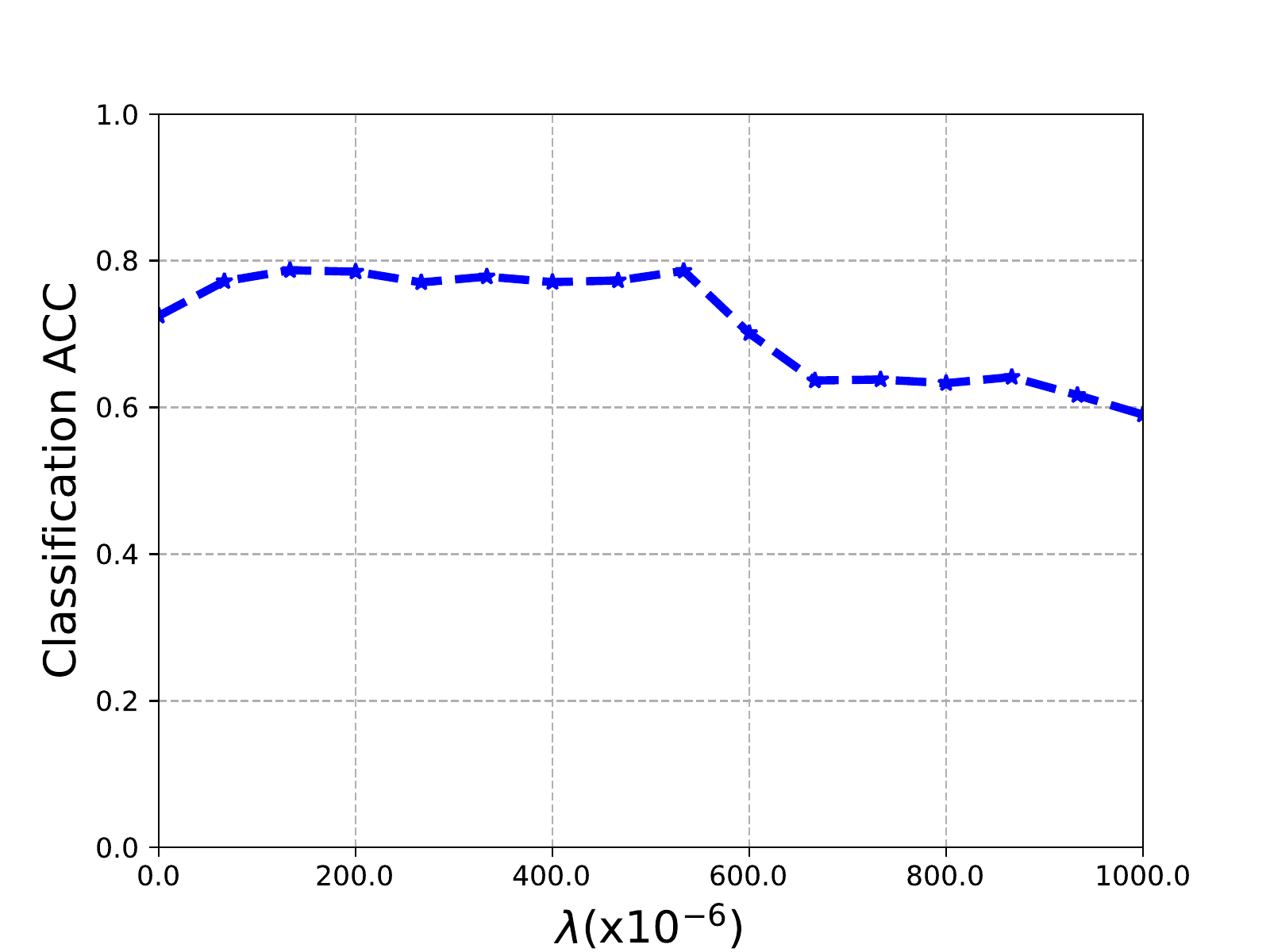}} 
 \caption{Parameter tuning on datasets.}
 \label{fig:par}
 \end{figure}

\begin{figure}[htb]
\centering
 \subfigure[Cora]{
 \includegraphics[width=0.31\columnwidth]{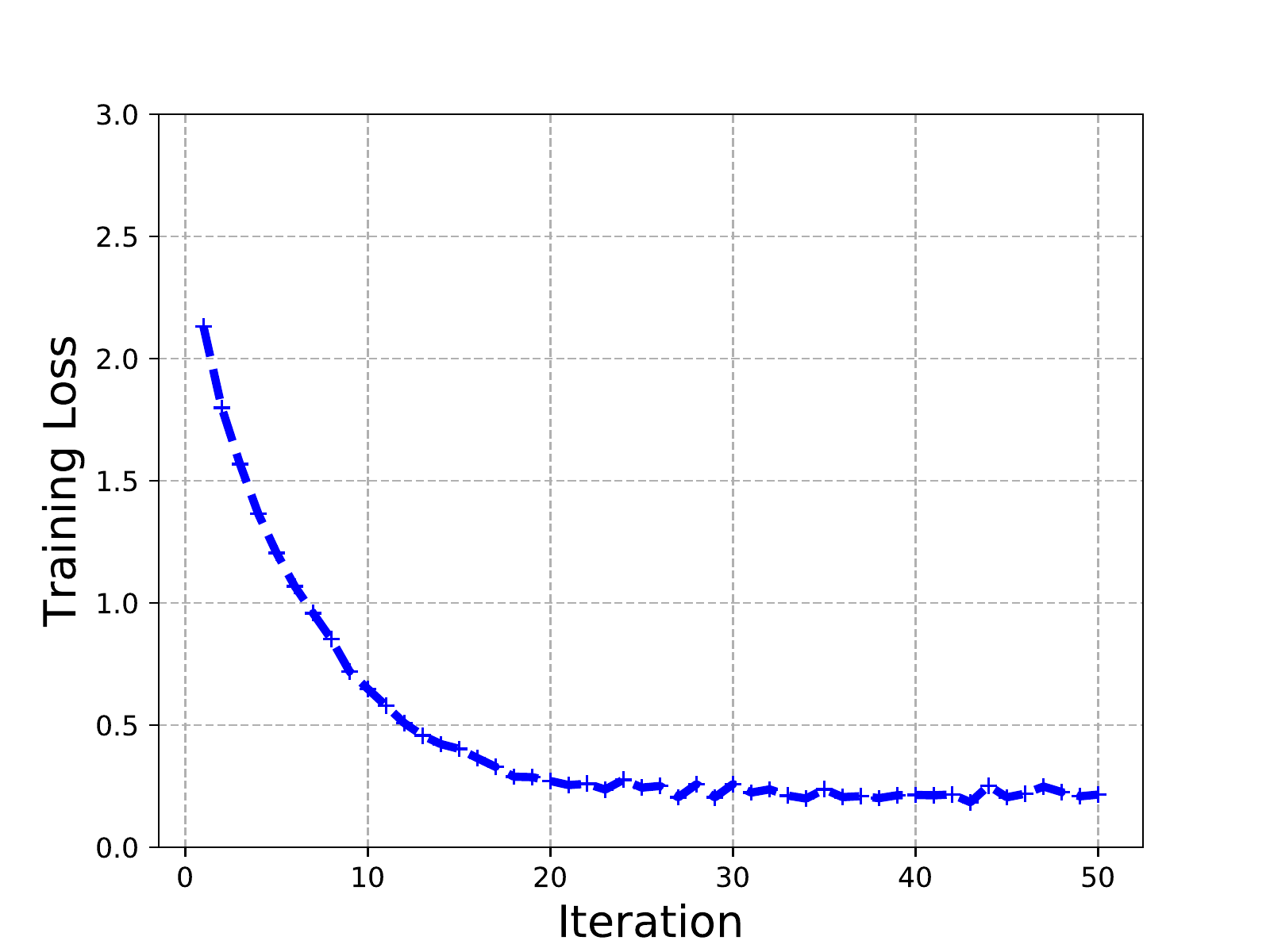}}
 \subfigure[Citeseer]{
 \includegraphics[width=0.31\columnwidth]{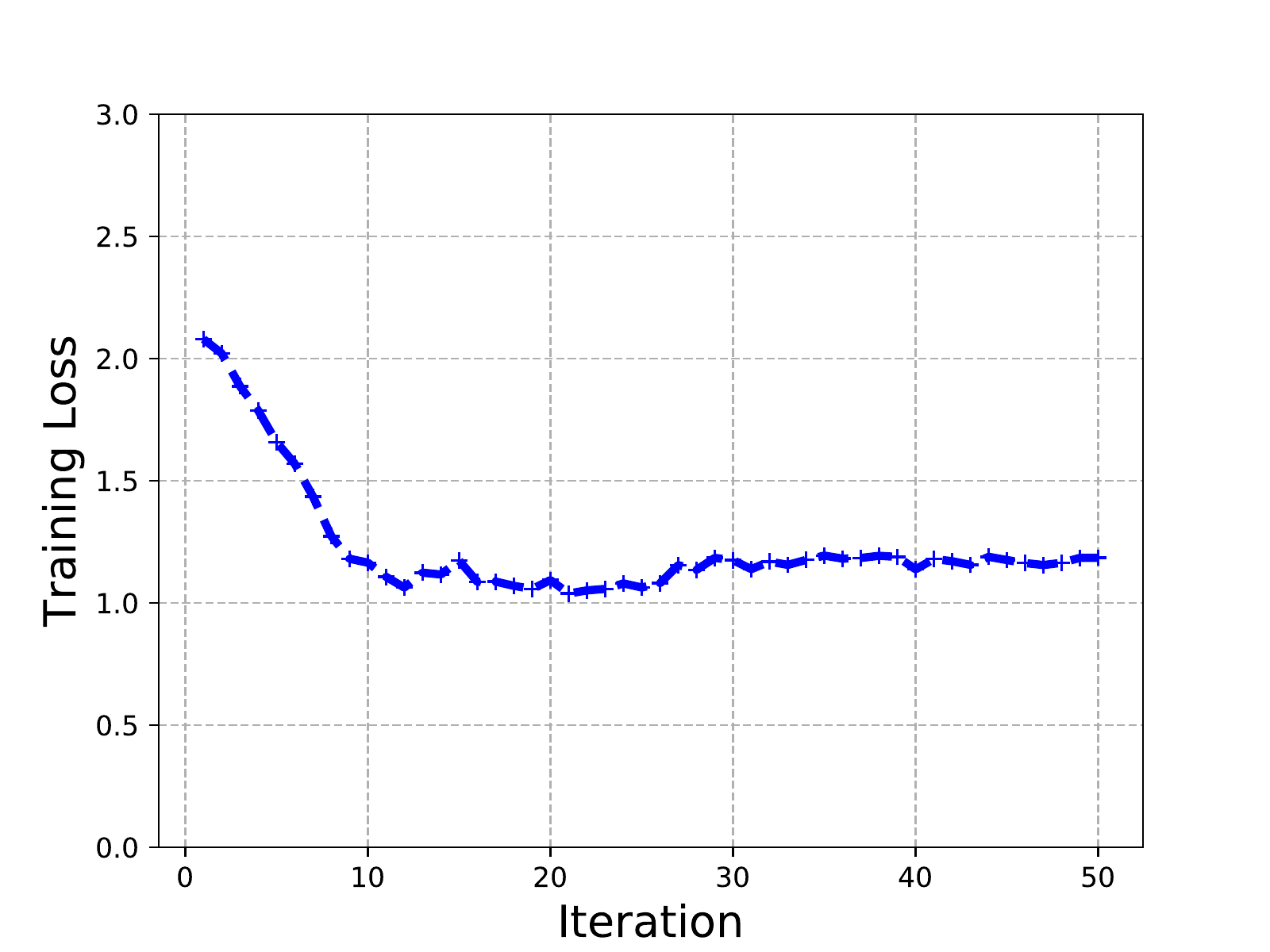}}
 \subfigure[Pubmed]{
 \includegraphics[width=0.31\columnwidth]{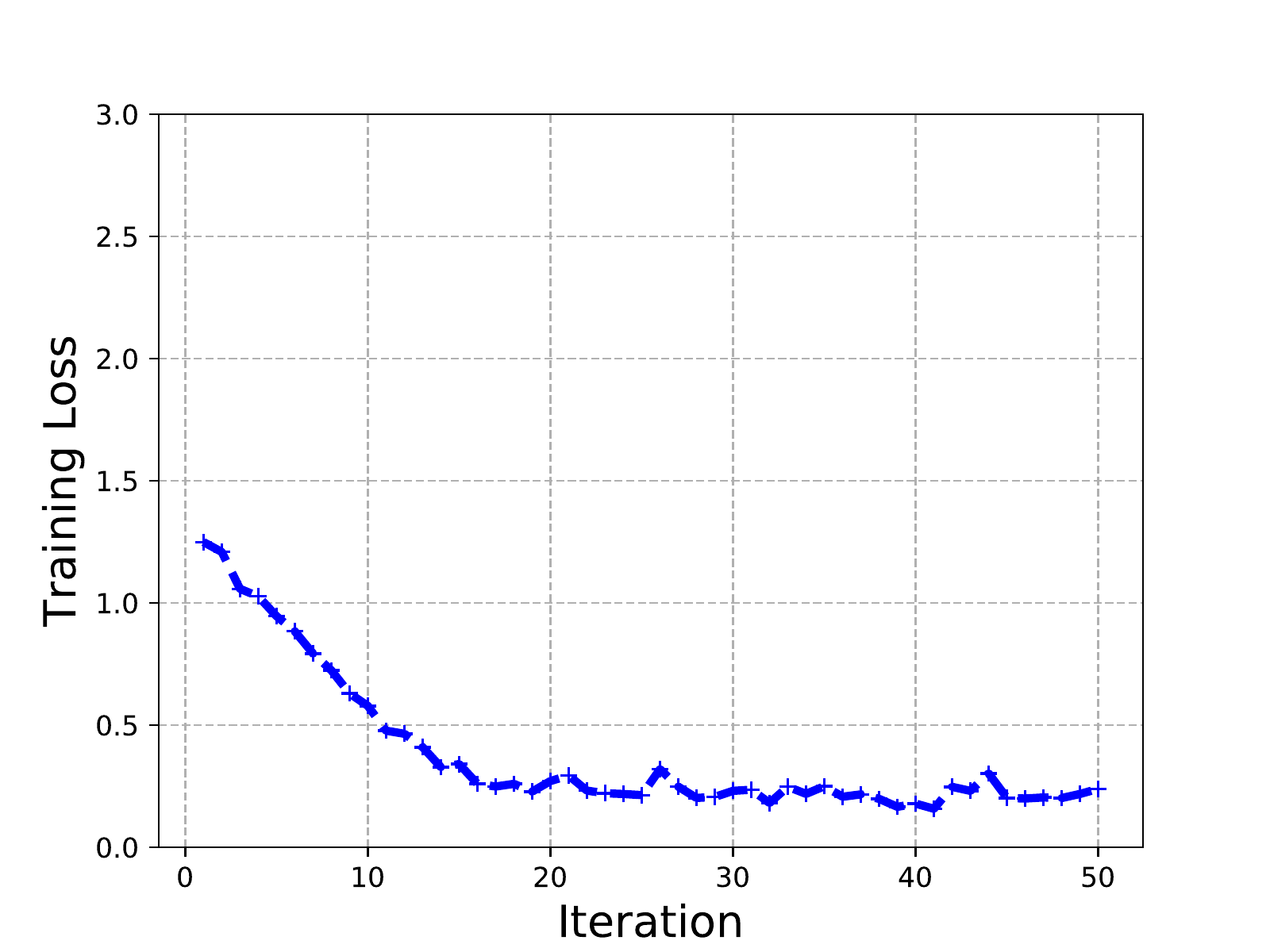}}
 \caption{Convergence results on datasets.}
 \label{fig:loss}
 \end{figure}

We show parameter adjustment and algorithm convergence on Cora as an example in Figure \ref{fig:par} and \ref{fig:loss}, respectively. From the Figure \ref{fig:par}, it can be seen that the parameter of independence term is relatively robust since the performance is stable while $\lambda$ is chosen in a wide range. Specifically, the promising performance can be expected when the parameter $\lambda$ is given in a range (e.g., $[10^{-5}, 10^{-6}]$). Yet the classification performance decreases dramatically when the value of $\lambda$ exceeds a upper limit (e.g., $1.25 \times 10^{-5}$), which demonstrate that it is not good to overemphasize the independence between latent factors.
From Figure \ref{fig:loss}, we can see that IPGDN converges within a small number of iterations, which empirically proves the efficiency of our model.

\section{Related work}

 
Inspired by the huge success of convolutional networks in the computer vision domain, recently there has been a large number of methods trying to apply convolution on non-Euclidean graph data. These approaches are under the umbrella of graph convolutional networks. Generally speaking, GCNs fall into two categories:spectral-based and spatial-based. 
Bruna et al. \cite{bruna2014spectral} propose the fist prominent work on GCNs, which designs a variant of graph convolution in the light of spectral graph theory. 
Subsequently, a large of increasing improvements, extensions, and approximations on spectral-based graph convolutional networks  \cite{defferrard2016convolutional,kipf2017semi-supervised,levie2018cayleynets} have been proposed. As spectral methods usually deal with the whole graph simultaneously and are hard to scale or parallel to large graphs, spatial-based graph convolutional networks have rapidly developed recently \cite{hamilton2017inductive,monti2017geometric,gao2018large}. These approaches directly formulate graph convolutions as aggregating feature information from neighbors in the graph domain. Furthermore, their computation can be largely decreased together with sampling strategies, which can be performed in a batch of nodes instead of the whole graph \cite{hamilton2017inductive,gao2018large}. Hence, they have the potential to improve efficiency.   

Another line of works related to ours is the disentangled representation learning. 
Early works that have demonstrated disentanglement in limited settings include \cite{desjardins2012disentangling},
and several prior researches have addressed the problem of disentanglement in supervised or semi-supervised settings \cite{kingma2014semi,kulkarni2015deep}.
Recently, disentangled representation learning has gained considerable attention, in particular in the field of image representation learning  \cite{alemi2016deep}. 
It is designed for learning representations to separate the explanatory factors of variations behind the data. These representations have proven to be more resilient to complex variants \cite{bengio2013representation} and can enhance enhanced generalization capabilities and improve the robustness of confrontational attacks \cite{alemi2016deep}. 
Furthermore, The disentangled representations are inherently easier to interpret and hence may be helpful for debugging and auditing \cite{doshi2017towards}. 
More recently, there are several recent works that constrain the form of this decomposition to capturing purely independent factors of variation in unsupervised generative models \cite{cao2015diversity,chen2018isolating,kim2018disentangling}. They typically evaluate the disentanglement using purpose-built, artificial, data and their generative factors are themselves independent by construction.
However, these approaches are mainly applied to traditional data but not to graph data.

\section{Conclusions}
In this paper, we propose a novel independence promoted graph disentangled networks for graph representations. In our approach, disentangled representation learning and independence measure among latent representations are integrated into a unified framework. Specifically, we present a HSIC scheme to regularize latent representations and enforce them to be independent.
The regularized module is jointly learned with a graph convolutional networks to produce the informative representations.
Experimental results demonstrate that our algorithm IPGDN outperform the state-of-the-arts in several graph node-related tasks.

\section{ Acknowledgments}
This work is supported in part by the National Natural Science Foundation of China (No. 61702296, 61972442, 61901297), Tianjin Science and Technology Major Projects and Engineering (No. 17ZXSCSY00060, 17ZXSCSY00090), the Program for Innovative Research Team in University of Tianjin (No. TD13-5034) and the 2019 CCF-Tencent Open Research Fund. 

\bibliographystyle{aaai}
\bibliography{ref}

\end{document}